\documentclass{article}

  \usepackage[preprint]{neurips_2026}

\usepackage[utf8]{inputenc} 
\usepackage[T1]{fontenc}    
\usepackage{amsmath} 
\usepackage{multirow} 
\usepackage{graphicx} 
\usepackage{hyperref}       
\usepackage{url}            
\usepackage{booktabs}       
\usepackage{amsfonts}       
\usepackage{nicefrac}       
\usepackage{microtype}      
\usepackage{xcolor}         

\title{Norm or Direction? Decoding Vision Mambas \\for High-Resolution Vision}

\author{%
  Jin Yu \\
  Center for Humanoid Research\\
  Korea Institute of Science and Technology\\
  \texttt{jinyu@kist.re.kr} \\
  \And
  Juyoun Park\thanks{corresponding author} \\
  Center for Humanoid Research\\
  Korea Institute of Science and Technology\\
  \texttt{juyounpark@kist.re.kr} \\
}

\begin{document}

\maketitle

\begin{abstract}
Vision Mamba models replace quadratic self-attention with linear complexity selective state space models (SSMs), emerging as efficient visual backbones. However, MambaOut demonstrates that a Gated CNN block can match or exceed VMamba on image classification, questioning the necessity of SSMs for vision. This raises a fundamental question: do VMamba and MambaOut encode visual information differently at the representation level? To investigate, we apply cross model centered kernel alignment (CKA) analysis and find that VMamba's final stage blocks form representations distinctly different from both MambaOut and its own preceding blocks. We therefore focus on the final block features, decomposing each spatial token into magnitude and direction. MambaOut concentrates class-discriminative information in high-norm foreground tokens that align with Grad-CAM attribution. VMamba, by contrast, produces high-norm tokens predominantly in background regions, misaligned with Grad-CAM, yet preserves discriminative signals primarily in token directions. These observations reveal that the two models rely on different encoding strategies. We connect this difference to high-resolution classification and semantic segmentation. VMamba distributes logit support broadly across object regions, whereas MambaOut relies on sparse dominant tokens, a strategy that becomes less stable as token counts grow. Under full fine-tuning for segmentation, VMamba consistently outperforms MambaOut. These results suggest that VMamba's advantage in dense prediction stems not merely from the SSM mechanism or sequence length, but from how semantic evidence is organized across token magnitude, direction. Ultimately, we conclude that token magnitude and directional structure serve as critical axes for improving visual backbones, particularly under dense supervision.
\end{abstract}

\section{Introduction}

Mamba~\citep{gu2024mamba} is an RNN-like sequence model that processes sequences with linear complexity through a selective scan mechanism, offering an efficient alternative to quadratic self-attention~\citep{vaswani2017attention}.
Recent Vision Mamba models~\citep{liu2024vmamba, yang2024plainmamba, huang2024localmamba, shaker2025groupmamba, hatamizadeh2025mambavision, pei2025efficientvmamba, zhu2024vim} adapt this principle to visual recognition. Among them, VMamba~\citep{liu2024vmamba} builds a hierarchical backbone via a 2D selective scan module (SS2D), achieving
competitive performance across image classification, object detection, and semantic segmentation.

However, MambaOut~\citep{yu2025mambaout} questions whether selective scan is strictly necessary for vision: replacing Mamba blocks with Gated CNN blocks~\citep{dauphin2017language, yu2024metanext} is sufficient to match or outperform VMamba on image classification. MambaOut attributes this to task properties, arguing that classification is neither a long-sequence nor an autoregressive task. Yet VMamba retains advantages on dense prediction benchmarks, where longer spatial sequences arise. This raises a fundamental question: do VMamba and MambaOut organize class-discriminative information in the same way, or do they rely on fundamentally different encoding strategies?

To investigate, we apply cross model centered kernel alignment (CKA)~\citep{cortes2012alcka, kornblith2019similarity} and find that VMamba's final stage blocks form representations that are distinct from both MambaOut and their own preceding blocks, exhibiting cross model divergence and greater sensitivity to supervision change. We trace this to a difference in
how each model encodes class-discriminative information. Decomposing spatial tokens into magnitude and direction, we find that MambaOut concentrates class information in high-norm foreground tokens that align with Grad-CAM~\citep{selvaraju2017gradcam} attribution, whereas VMamba produces high-norm tokens predominantly in background regions and preserves discriminative signals primarily in token directions. We then connect this representational difference to task performance. Token replacement attribution reveals that VMamba distributes logit support broadly across object regions, while MambaOut concentrates it in fewer dominant tokens. This distributional difference is consistent with  VMamba's growing advantage as token count increases at higher
resolutions. Under dense supervision, VMamba's direction-based encoding reorganizes more readily into spatially selective representations, contributing to its stronger performance under full fine-tuning for segmentation. The unit-token decoder test further confirms that VMamba's dense prediction remains largely intact when token magnitudes are removed, while MambaOut degrades sharply.

Taken together, these findings suggest that VMamba's advantage in dense prediction stems not merely from long-range dependency modeling, but from how discriminative information is distributed across token directions. Our results identify token magnitude and direction as key axes for designing high-resolution visual backbones, and point to foreground-aligned magnitude regularization and angular regularization of token directions as concrete directions for improving performance in dense and high-resolution settings.

\section{Related Work}
\label{sec:related}
\paragraph{Generic visual backbones.}
Convolutional neural networks (CNNs)~\citep{krizhevsky2012alexnet, Si2015vggnet, he2016resnet, xie2017resnext} have long served as the dominant backbone for visual recognition. Vision Transformers (ViT)~\citep{dosovitskiy2021vit, touvron2021deit, yuan2021t2tvit, han2021tit} later demonstrated superiority over CNNs in classification. To better support multi-scale visual recognition, subsequent work introduced hierarchical designs~\citep{liu2021swin, yang2021focal, wang2021pyramid, dong2022cswin, ding2022davit, zhang2023hivit}. However, the quadratic complexity of dot-product attention~\citep{vaswani2017attention} poses fundamental challenges in processing long sequences~\citep{tay2022survey}, which becomes particularly problematic for high-resolution visual inputs. Linear attention~\citep{katharopoulos2020linearattention} offers inherent $O(N)$ complexity and has since been widely explored~\citep{choromanski2021rethinking, zhen2022cosformer, xiong2021nystromformer, peng2023rwkv}. Despite their efficiency, linear attention methods consistently underperform softmax attention, limiting their practical adoption.

\paragraph{State Space Models (SSMs) for visual recognition.}
SSMs provide an alternative path to linear complexity modeling by capturing long-range dependencies through structured state transitions. 
Early approaches such as S4~\citep{gu2022efficiently} parameterize linear time-invariant (LTI) systems with static, input-independent kernels and were originally designed for 1D sequence modeling. S4ND~\citep{nguyen2022s4nd} extends this formulation to images via an outer-product construction while retaining the same static limitation. Mamba~\citep{gu2024mamba} addresses this limitation by introducing input-dependent selective scanning, improving adaptability without sacrificing linear complexity. 
Recent efforts adapt Mamba to vision by applying sequence modeling to flattened 2D feature maps~\citep{liu2024vmamba, yang2024plainmamba, huang2024localmamba, shaker2025groupmamba, pei2025efficientvmamba}, often incorporating multi-directional or windowed scanning to better capture spatial structure. Among these, VMamba~\citep{liu2024vmamba} proposes a 2D selective scan (SS2D) over four directions. 
In contrast, MambaOut~\citep{yu2025mambaout} questions the necessity of SSMs for image classification, while leaving open their role in dense prediction. Motivated by this, we study how VMamba and MambaOut differ in their internal representations and how these differences relate to performance on high-resolution dense prediction tasks.

\section{Background}
\label{sec:background}

\paragraph{State Space Models (SSMs).}
Originating from the Kalman filter~\citep{kalman1960new}, structured SSMs such as S4~\citep{gu2022efficiently} are linear time-invariant (LTI) systems that map the input signal \(u(t)\in\mathbb{R}\) to the output response \(y(t)\in\mathbb{R}\) via the hidden state \(h(t)\in\mathbb{R}^N\), expressed as linear Ordinary Differential equations (ODEs):

\begin{equation}
    h'(t) = Ah(t) + Bu(t),\ y(t) = Ch(t) + Du(t)
\end{equation}

where \(A\in\mathbb{R}^{N\times N}, B\in\mathbb{R}^{N\times 1}, C\in\mathbb{R}^{1\times N}\), and \(D\in\mathbb{R}^{1\times 1}\) are weighting parameters. Mamba~\citep{gu2024mamba} introduces a selective state space model (S6) with an explicit discretization based on the Zero-Order Hold (ZOH) method:

\begin{equation}
\begin{aligned}
    \bar{A} &= exp(\Delta A), \bar{B}=(\Delta A)^{-1}(exp(\Delta A)-I)\cdot\Delta B \\
    h_t &= \bar{A}h_{t-1}+\bar{B}u_t,\ y_t = Ch_t + Du_t
\end{aligned}
\end{equation}

\(\Delta\) is a timescale parameter that controls the discretization interval. In S6, \(B\), \(C\), and \(\Delta\) are input-dependent, enabling content-aware sequence modeling while maintaining linear computational complexity.

Adapting S6 to vision requires bridging 1D sequential scanning and 2D spatial structure. A common approach flattens the 2D feature map into a 1D sequence prior to applying S6~\citep{zhu2024vim, liu2024vmamba, yang2024plainmamba, huang2024localmamba, shaker2025groupmamba}. VMamba~\citep{liu2024vmamba} follows this approach, scanning the flattened sequence along four complementary directions via the proposed SS2D module.

\paragraph{2D Selective Scan (SS2D).}
Images do not have a natural one-dimensional order. SS2D addresses this by converting a 2D feature map into multiple 1D sequences. Given an input feature map, SS2D scans it in four directions: top-to-bottom, bottom-to-top, left-to-right, and right-to-left. Each sequence is processed by an S6 block~\citep{gu2024mamba}, and the resulting sequences are reshaped and merged back into the original 2D spatial layout. Through these complementary scan directions, each spatial token can aggregate long-range context from different parts of the image.

\begin{figure}[t]
  \centering
  \includegraphics[width=1.0\linewidth]{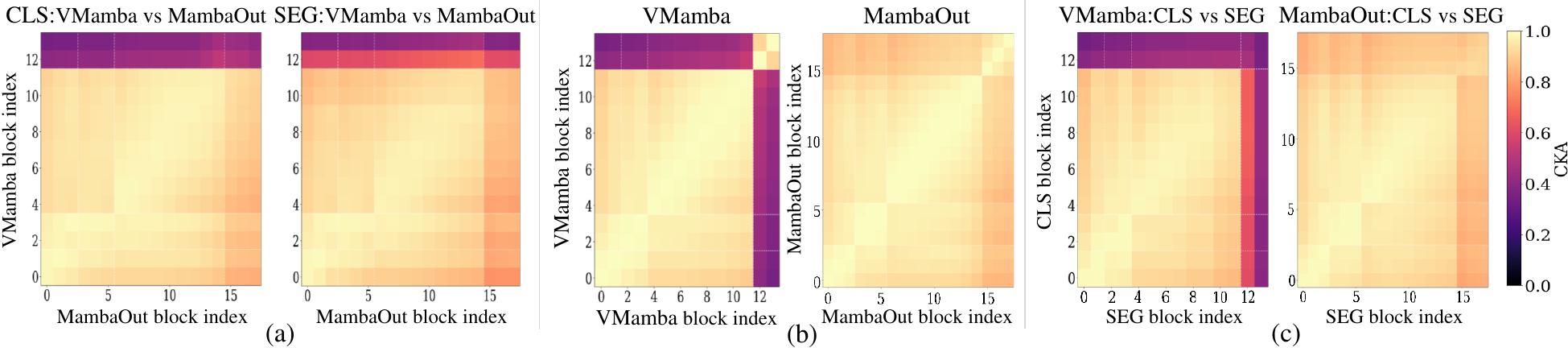}
      \caption{CKA similarity heatmaps between VMamba-T and MambaOut-T. (a) Cross model CKA for classification-pretrained (CLS) and segmentation fine-tuned (SEG) backbone, and (b) within model CKA across all blocks of VMamba and MambaOut. (c) Within model CKA between CLS and SEG representations.}
  \label{fig:cka}
\end{figure}

\paragraph{VMamba and MambaOut.}
VMamba~\citep{liu2024vmamba} and MambaOut~\citep{yu2025mambaout} adopt a hierarchical four-stage backbone similar to ResNet~\citep{he2016resnet}, where feature resolution is gradually reduced across stages while the channel dimension increases. In both models, the final feature map is aggregated via global average pooling (GAP)~\citep{lin2013nin, szegedy2015googlenet} before being passed to a classifier head. VMamba's Visual State Space (VSS) block follows the architecture of vanilla Transformer block~\citep{vaswani2017attention}, where the SS2D module replaces self-attention as the token mixer, followed by an FFN, with residual connections around each. 
MambaOut instead uses a Gated CNN block~\citep{dauphin2017language, yu2024metanext}, where a $7{\times}7$ depthwise convolution serves as the token mixer within a gating mechanism, following ConvNeXt~\citep{liu2022convnet}. MambaOut argues that SSM is unnecessary for standard image classification because classification is neither a long-sequence nor an autoregressive task. However, it leaves open the possibility that SSM remains useful for dense prediction, where the token sequence is much longer. This motivates our representation analysis: if the two models behave differently across tasks, their internal representations may also differ in a systematic way.

\section{Representation Analysis}
\label{sec:rep}
We compare how VMamba and MambaOut construct visual representations using the Tiny models (VMamba-T and MambaOut-T) throughout all experiments. We begin by identifying where representational differences arise across stages, then focus on how the final stage features encode class-discriminative information.

\subsection{Block-level Representation Similarity}
\paragraph{Centered Kernel Alignment (CKA).}
\label{sec:cka}
We apply CKA~\citep{cortes2012alcka, kornblith2019similarity} to compare block-level representations across all stages of VMamba and MambaOut. CKA measures the similarity between two sets of representations evaluated on the same inputs, with a score of 1 indicating identical representations up to orthogonal transformation and 0 indicating no similarity. We construct three types of heatmap: cross model comparisons between VMamba and MambaOut, within model comparisons across all block pairs in each model, and within model comparisons between classification pretrained and segmentation fine-tuned backbones. Implementation details are provided in Appendix~\ref{app:cka}.

\begin{figure}[t]
  \centering
  \includegraphics[width=0.8\linewidth]{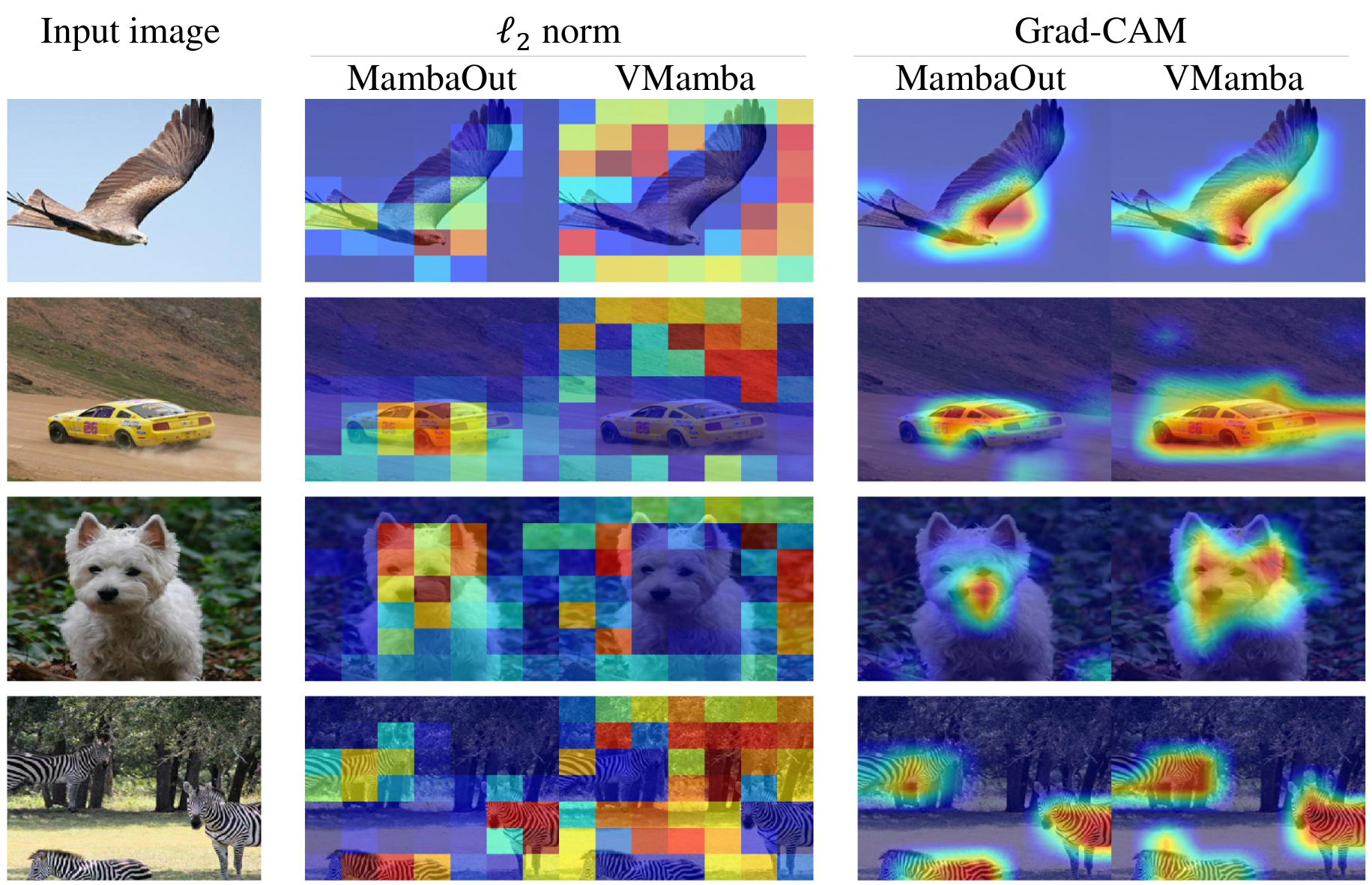}
  \caption{Visual comparison of token \(\ell_2\) norm and Grad-CAM heatmaps from the final block outputs of VMamba and MambaOut.}
  \label{fig:l2-grad}
\end{figure}

\paragraph{Observation.}
As shown in Figure~\ref{fig:cka}(a), earlier stages produce high CKA scores across the two models, indicating similar representations. 
However, VMamba’s final stage blocks show low CKA scores not only with MambaOut but also with their own preceding blocks (Figure~\ref{fig:cka}(b)), whereas MambaOut maintains consistently similar representations across layers.
Figure~\ref{fig:cka}(c) further shows that VMamba's classification and segmentation representations exhibit low CKA scores at the final stage, whereas MambaOut maintains high CKA scores throughout. 
These results indicate that VMamba's final stage forms qualitatively different representations, exhibiting both cross model divergence and sensitivity to supervision changes. This motivates a closer analysis of what they encode.

\subsection{Where do high activations appear?}

\paragraph{\(\ell_2\) norm and Grad-CAM.}
To investigate what distinguishes VMamba's final stage representations, we examine the spatial distribution of token activations. A high token norm indicates that the corresponding spatial location has a large activation magnitude. We compare this norm map with Grad-CAM~\citep{selvaraju2017gradcam}, using the predicted class logit as the target. Grad-CAM identifies which spatial regions contribute the most to the classification decision.

\begin{figure}[t]
  \centering
  \includegraphics[width=0.99\linewidth]{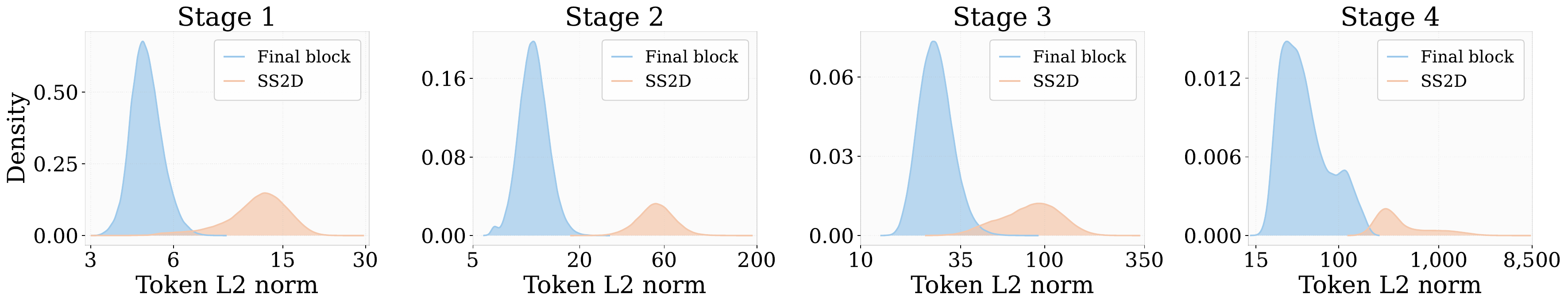}
  \caption{Token norm distributions across stages and block components.}
  \label{fig:norm-dis}
\end{figure}

\begin{figure}[t]
  \centering
  \includegraphics[width=1.0\linewidth]{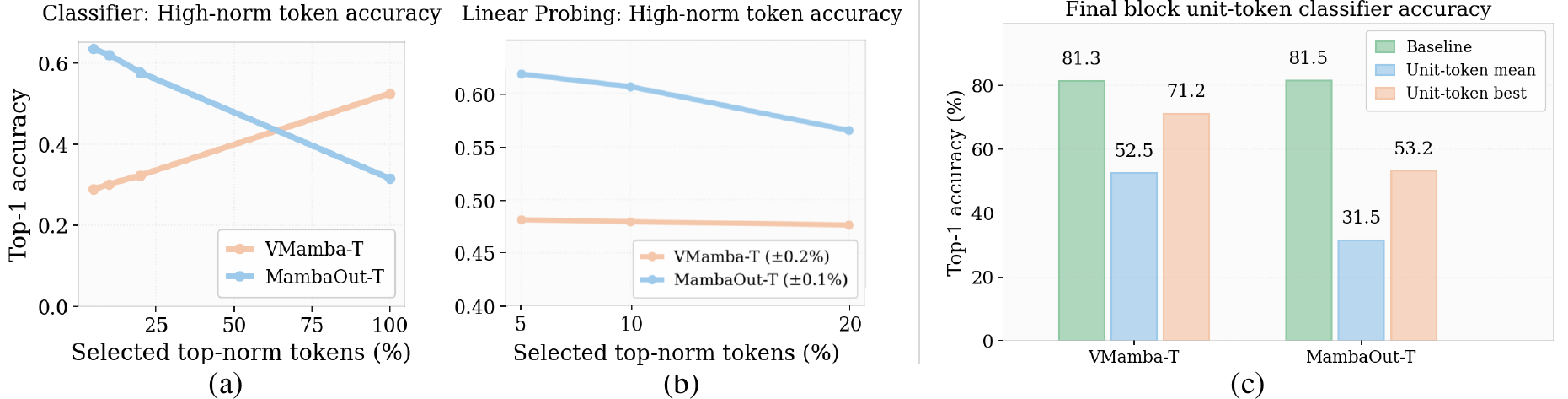}
  \caption{(a) High-norm token accuracy, (b) high-norm linear probing, and (c) unit-vector evaluation for VMamba and MambaOut.}
  \label{fig:high-norm}
\end{figure}

\paragraph{Observation.}
Figure~\ref{fig:l2-grad} shows a clear difference between the two models. In  MambaOut, high-norm tokens are concentrated on foreground objects, and the Grad-CAM heatmap is also foreground-focused. Thus, token magnitude and classification attribution are spatially aligned.
VMamba behaves differently. Its high-norm tokens often appear in background regions, while Grad-CAM remains concentrated on the foreground object. This spatial mismatch raises a direct question: do VMamba's high-norm background tokens actually carry class-discriminative information?
Previous works~\citep{darcet2024vit-reg, wang2025mamba-reg} on ViT~\citep{touvron2022deit-iii, ilharco2021openclip, oquab2024dinov2} and Vision Mamba~\citep{zhu2024vim} have reported a similar background concentration of high-norm tokens, characterizing them as global information repositories with a clearly bimodal norm distribution. VMamba, however, presents a subtly different picture. Rather than a discrete outlier population, its norm distribution is smooth and unimodal (Figure~\ref{fig:norm-dis}). This pattern suggests that VMamba's background tokens may not arise from the same outlier mechanism reported in prior work, and we investigate whether they nonetheless encode class-discriminative content in the following subsection.

\subsection{Do high-norm tokens carry class information?}

The previous observation raises a direct question: are VMamba's high-norm background tokens actually informative for classification? Prior work~\citep{darcet2024vit-reg, wang2025mamba-reg} tests similar outlier tokens by using them for image classification. If high-norm tokens are the primary carriers of class-discriminative information, a classifier should be able to predict the image label using only those tokens. We evaluate this under two protocols: feeding high-norm tokens directly to the pretrained classification head, and training a linear probe via ridge regression, with implementation details in Appendix~\ref{app:high_norm_test} and~\ref{app:linear_prob}.

\paragraph{High norm test.}
\label{sec:high_norm_test}
Figure~\ref{fig:high-norm}(a) reports the average accuracy obtained from the top \(5\%\), \(10\%\) and \(20\%\) high-norm tokens. In MambaOut, accuracy is highest when using only the most high-norm tokens and decreases as more tokens are included. This indicates that class-discriminative evidence is concentrated in a small number of high-norm tokens, suggesting that MambaOut uses token magnitude as a strong cue for class-discriminative information.
VMamba shows the opposite trend. Accuracy improves as more tokens are included. This means that the high-norm tokens are not necessarily the most informative ones. This suggests that VMamba does not use high-norm tokens as its primary class-information carriers.

\paragraph{High norm linear probing.}
\label{sec:high_norm_linear_probing}
Figure~\ref{fig:high-norm}(b) shows a similar pattern under linear probing. MambaOut performs best with fewer high-norm tokens, again suggesting that class information is concentrated in a small high magnitude subset. VMamba shows relatively small variation across token ratios. This suggests that its class information is less concentrated in high-norm tokens and more distributed across spatial positions.

\subsection{VMamba preserves information in token directions}

If VMamba does not primarily encode class information in token magnitude, where is the information preserved? A token vector can be decomposed into magnitude and direction, and we focus on the directional component. Specifically, we normalize each token to unit norm and evaluate individual normalized tokens using the pretrained classification head. This removes magnitude information and leaves only direction. If classification accuracy remains high after normalization, then class-discriminative information must be preserved in the angular component of the representation. Implementation details are provided in Appendix~\ref{app:unit_norm_test}.

\begin{figure}[t]
  \centering
  \includegraphics[width=0.7\linewidth]{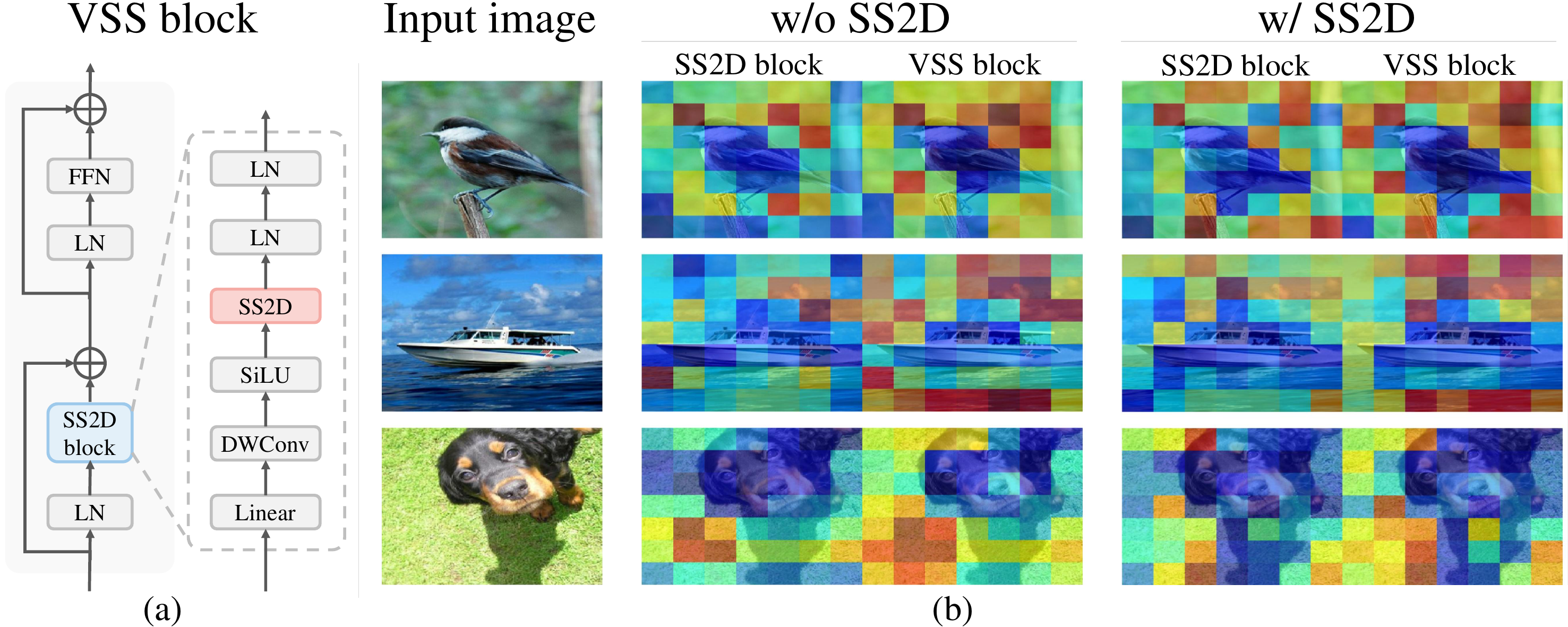}
  \caption{(a) Architecture of the VMamba's VSS block, where the SS2D block applies SS2D as its core scanning operation. (b) Token $\ell_2$ norm heatmaps of two intermediate outputs (from the SS2D block and VSS block) under two conditions (with and without SS2D). The SS2D block without SS2D refers to the block with only the SS2D module removed.}
  \label{fig:wossm}
\end{figure}

\paragraph{Unit-vector test.}
\label{sec:unit_vector_test1}
Figure~\ref{fig:high-norm}(c) shows that MambaOut suffers a substantial drop after token-wise normalization. Its average single-token accuracy becomes \(31.5\%\), indicating that much of its discriminative signal depends on magnitude. VMamba remains much stronger after normalization. Its best single-token accuracy reaches \(71.2\%\), and its average accuracy is \(52.5\%\), which is \(21\%\) higher than MambaOut.
These results suggest that VMamba and MambaOut use different encoding strategies. MambaOut relies more heavily on token magnitude. VMamba, by contrast, preserves class-discriminative information more strongly in token directions. 
This suggests a representation where information is preserved in token directions and distributed across spatial tokens, enabling both aggregation across tokens and adaptability under dense supervision. We test these implications in Section~\ref{sec:dense}. Before that, we investigate what gives rise to this representation.

\subsection{What gives rise to background high-norm?}

Having established that VMamba's high-norm tokens are concentrated in background regions, we ask what gives rise to this pattern. A natural hypothesis is that this pattern is caused by the SS2D, since SS2D is the main selective scan mechanism in VMamba. However, our observations suggest that background high-norm is not solely an SSM-specific effect. As shown in Figure~\ref{fig:wossm}(b), the background high-norm pattern still appears when SS2D is removed from VMamba and the model is retrained. These results suggest that the phenomenon is not solely attributable to the selective scan mechanism.

These results characterize VMamba's representation. High-norm tokens are not necessarily aligned with semantic foreground evidence, while \textit{class-discriminative information remains preserved in token directions} across spatial locations. This is consistent with the observations in Section~\ref{sec:cka}, where VMamba differs from MambaOut and shows larger changes under different supervision. We next examine how this representation behaves when the task requires longer token sequences or dense spatial predictions.

\section{High-Resolution Dense Prediction}
\label{sec:dense}
We now relate the representation analysis to high-resolution recognition. Our goal is to test whether the representation differences identified in Section~\ref{sec:rep} translate into performance differences when the task departs from standard \(224\times224\) ImageNet classification~\citep{deng2009imagenet}. We focus on two implications from Section~\ref{sec:unit_vector_test1}: stable evidence aggregation as the number of tokens increases, and adaptability under dense supervision. We consider three settings designed to disentangle distinct factors: generalization robustness, sequence length, and dense spatial supervision.

\subsection{High-resolution classification}
\label{sec:high_res_cls}
We ask whether VMamba's advantage in dense prediction stems from sequence length or whether it reflects a more fundamental representational difference. To separate these two factors, we evaluate both models at \(768\times768\) resolution, where the number of spatial tokens is approximately 11 times larger than at \(224\times224\), while the task remains classification. 
Since segmentation already constitutes a long-sequence dense prediction setting analyzed separately below, we apply high-resolution fine-tuning only to the classification task. We fine-tune the \(224\times224\)-pretrained VMamba and MambaOut models on ImageNet at \(768\times768\) resolution for 30 epochs. We also evaluate on ObjectNet~\citep{barbu2019objectnet}, a robustness benchmark in which test images are collected under controlled variation in viewpoint, rotation, and background, following the evaluation protocol of~\citep{taori2020measuring, radford2021learning}. Since ObjectNet has no training set, evaluation is zero-shot. We include this as a robustness check, testing whether MambaOut's advantage at \(224\times224\) persists under changes in viewpoint, rotation, and background. 

Table~\ref{tab:cls} shows that MambaOut outperforms VMamba at \(224\times224\), with \(81.5\%\) vs. \(81.3\%\) top-1 accuracy on ImageNet-1k. The same trend holds on ObjectNet zero-shot evaluation, where MambaOut achieves \(46.1\%\) compared with VMamba's \(45.4\%\), suggesting that its standard-resolution advantage is not confined to the ImageNet validation distribution. After high-resolution fine-tuning at \(768\times768\), however, VMamba obtains \(83.4\%\), slightly surpassing MambaOut at \(83.3\%\). At this resolution, the number of spatial tokens is approximately 11 times larger. To better understand this change, we examine how the target logit is supported by individual final stage spatial tokens at both resolutions.

\begin{table}[t]
  \centering
  \caption{Classification comparison. MambaOut performs slightly better at standard resolution, whereas VMamba slightly surpasses MambaOut after high-resolution fine-tuning.}
  \label{tab:cls}
  \small
  \begin{tabular}{llllcc}
    \toprule
    Task & Dataset & Res. & Setting & VMamba-Tiny & MambaOut-Tiny \\
    \midrule
    \multirow{3}{*}{Classification}
      & ImageNet-1k & 224 & Pretrained & 81.3 & \textbf{81.5} \\
      & ObjectNet & 224 & Zero-shot & 45.4 & \textbf{46.1} \\
      & ImageNet-1k & 768 & Fine-tuning & \textbf{83.4} & 83.3 \\
    \bottomrule
  \end{tabular}
\end{table}

\begin{figure}[t]
  \centering
  \includegraphics[width=0.9\linewidth]{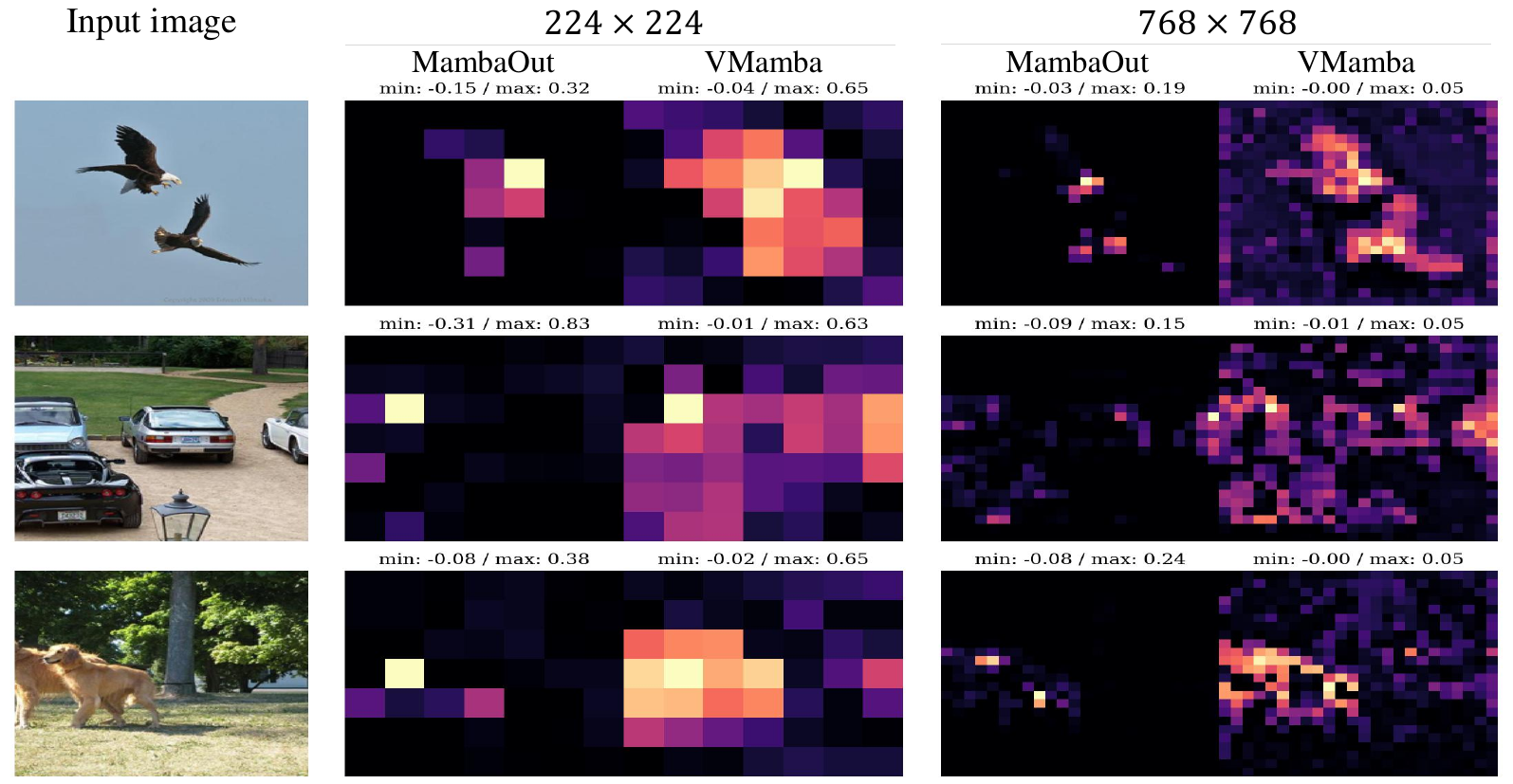}
  \caption{Token replacement attribution at \(224\times224\) and \(768\times768\) resolution for VMamba and MambaOut. For each final block spatial token, we replace only that token with the image-wise mean token and measure the drop in the target class logit.}
  \label{fig:token-replace}
\end{figure}

\paragraph{Token replacement attribution.}
\label{sec:token_replacement1}
We further analyze evidence aggregation using a token-level perturbation analysis. Inspired by perturbation-based visual explanation methods~\citep{zeiler2014visualizing, fong2017interpretable}, we adapt token-level perturbation to measure evidence aggregation at the final stage. Given the feature map $X=\{x_i\}_{i=1}^{N}$, we replace each token $x_i$ with the image-wise mean $\bar{x}=\frac{1}{N}\sum_{j=1}^{N}x_j$ and measure the resulting drop in the target class logit. We define the token replacement score as
\begin{equation}
    e_i = f_c(X) - f_c\!\bigl(X^{(i \leftarrow \bar{x})}\bigr),
\end{equation}
where $f_c(\cdot)$ denotes the target class logit. A larger $e_i$ indicates that token $i$ provides stronger positive support for the prediction. Implementation details are provided in Appendix~\ref{app:token_replacement}.

In Figure~\ref{fig:token-replace}, brighter regions indicate larger positive logit drops. The min/max values shown above each heatmap denote the minimum and maximum token replacement scores within that heatmap. This figure shows that VMamba and MambaOut exhibit different evidence aggregation patterns. At both \(224\times224\) and \(768\times768\), VMamba shows logit-supporting evidence over broader object regions, whereas MambaOut concentrates evidence in fewer object locations. This suggests that MambaOut relies on sharper but sparser evidence peaks, while VMamba uses a broader spatial support. This observation provides a representation-level correlate of the high-resolution trend. It suggests that distributing information across token directions enables more effective aggregation as the number of tokens increases. At \(224\times224\), concentrating evidence in a few object tokens can be sufficient for image-level classification, consistent with MambaOut's advantage. At \(768\times768\), however, the final feature map contains more spatial tokens. In this setting, relying on a small set of dominant evidence tokens may make aggregation more sensitive to spatial sampling and resolution changes, whereas VMamba's broader evidence support may provide a more stable basis for global classification.

\begin{table}[t]
  \centering
  \caption{Segmentation comparison. MambaOut is stronger with a frozen backbone, whereas VMamba is stronger after full fine-tuning.}
  \label{tab:seg}
  \small
  \begin{tabular}{llllcc}
    \toprule
    Task & Dataset & Res. & Setting & VMamba-Tiny & MambaOut-Tiny \\
    \midrule
    \multirow{4}{*}{Segmentation}
      & ADE20K & 512 & Backbone frozen & 40.2 & \textbf{40.7} \\
      & ADE20K & 512 & Full fine-tuning & \textbf{47.9} & 47.1 \\
      & ADE20K & 512 & Unit-token input & \textbf{47.5} & 41.5 \\
      & Pascal VOC 2012 & 512 & Full fine-tuning & \textbf{81.6} & 79.2 \\
    \bottomrule
  \end{tabular}
\end{table}

\begin{figure}[t]
  \centering
  \includegraphics[width=0.99\linewidth]{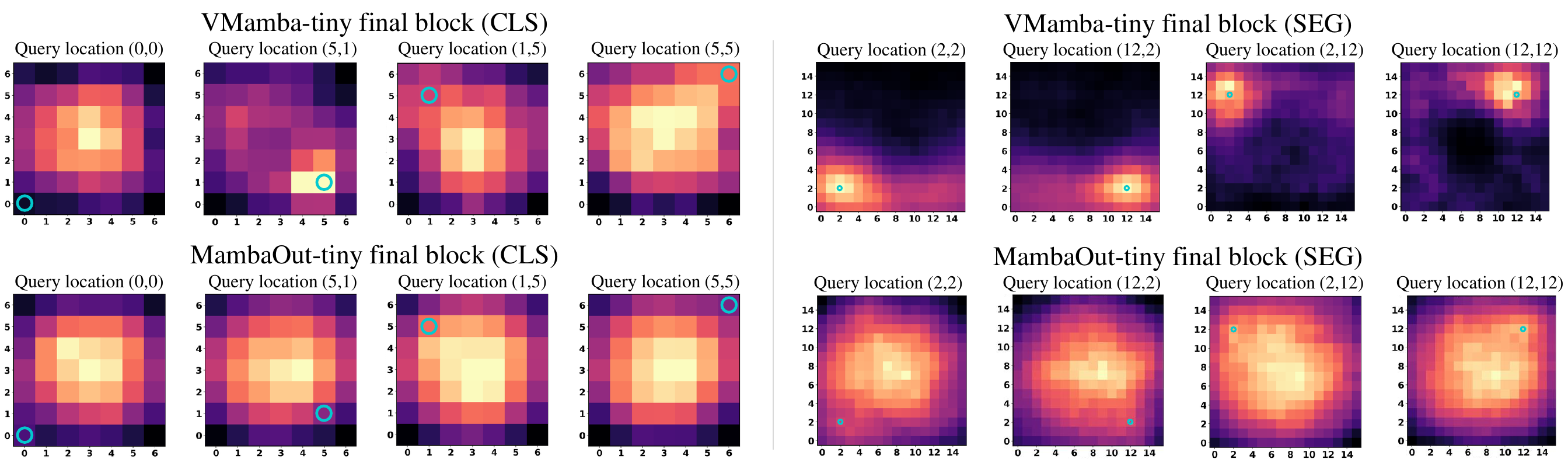}
  \caption{CKA-based spatial localization comparison between classification-pretrained and segmentation fine-tuned VMamba and MambaOut. (Green circle: query location)}
  \label{fig:spatial}
\end{figure}

\subsection{Semantic segmentation}
Semantic segmentation differs from classification in a fundamental way. Classification predicts one label for the whole image, so only image-level evidence is directly supervised. Segmentation predicts a label for every spatial location. Therefore, background regions are not merely context; they are part of the prediction target.
This distinction changes how we interpret VMamba's background high-norm tokens. In classification, they appear misaligned with class-discriminative attribution, with Grad-CAM focusing on foreground objects and high-norm tokens in the background. In segmentation, however, every spatial location is supervised, making this misalignment less consequential. We further investigate whether dense supervision alters the spatial structure of the learned representations.

\paragraph{Frozen backbone.}
When the backbone is frozen, MambaOut outperforms VMamba on ADE20K, achieving \(40.7\) mIoU compared with VMamba's \(40.2\) mIoU (first row of Table~\ref{tab:seg}). This is consistent with MambaOut's magnitude-aligned foreground evidence: a newly trained segmentation head can exploit these strong spatial magnitude cues without updating the backbone.

\paragraph{Full fine-tuning.}
When the entire backbone is fine-tuned, VMamba outperforms MambaOut on ADE20K, achieving \(47.9\) mIoU compared with \(47.1\), and shows a larger margin on Pascal VOC 2012~\citep{everingham2010pascal}, where VMamba reaches \(81.6\) mIoU against MambaOut's \(79.2\) (second and fourth row of Table~\ref{tab:seg}). This pattern is also reflected in the CKA analysis (Figure~\ref{fig:cka}). VMamba's representations change substantially at the final stage under dense supervision, suggesting that backbone updates induce a representational structure better suited to segmentation. This aligns with the observed performance reversal. It also raises a natural question: does VMamba's advantage under fine-tuning primarily arise from its token directions, or does it also depend on magnitude information?

\paragraph{Unit-token decoder test.}
\label{sec:unit_token_decoder}

To test whether dense prediction can be supported by token direction alone, we replace each stage feature passed to the decoder with its $\ell_2$-normalized direction at a preserved feature scale. Under this modification, VMamba drops only slightly on ADE20K~\citep{zhou2017ade}, from \(47.9\) to \(47.5\) mIoU. In contrast, MambaOut declines sharply from \(47.1\) to \(41.5\) mIoU (third row of Table~\ref{tab:seg}). This suggests that VMamba's dense prediction representation remains largely usable when raw token magnitude is removed, whereas MambaOut tends to rely more on magnitude information. This does not imply that VMamba ignores magnitude, but it indicates that its token directions are more sufficient for dense prediction than those of MambaOut.

\paragraph{Spatial localization by CKA.}
\label{sec:spatial_cka}
We examine whether dense supervision changes the spatial structure of the learned representations using a CKA-based spatial localization analysis~\citep{raghu2021vision}, 
evaluating both classification-pretrained and segmentation fine-tuned backbones on ImageNet-1K validation images. Details are in Appendix~\ref{app:cka}. As shown in Figure~\ref{fig:spatial}, classification-pretrained models show weak spatial localization, whereas segmentation fine-tuned VMamba exhibits markedly clearer spatial selectivity. MambaOut shows little change after fine-tuning, consistent with its magnitude-concentrated encoding being less amenable to spatial reorganization. These results indicate that VMamba's direction-based encoding is more readily reshaped by dense supervision into spatially selective representations, consistent with the within model CKA shift observed in Section~\ref{sec:cka}.

The CKA analysis and unit-token decoder test suggest two complementary reasons for VMamba's fine-tuning advantage. Under dense supervision, VMamba develops clearer spatial selectivity, and its token directions retain useful semantic and spatial information even when magnitude is removed. Together, these results indicate that directional encoding yields a more adaptable representation once the backbone is updated under dense supervision. Overall, the results suggest that VMamba's advantage is not explained solely by the presence of SSM. Its representation appears better suited to adaptation under dense spatial supervision. MambaOut appears easier to exploit with a frozen backbone, possibly because its magnitude-aligned foreground evidence provides accessible spatial cues for the segmentation head. \textit{VMamba, in contrast, becomes stronger under full fine-tuning, where dense supervision can reshape spatial selectivity and exploit direction-preserved semantic evidence.}

\section{Discussion}

VMamba and MambaOut encode class-discriminative information through
distinctly different strategies. MambaOut concentrates on a
small set of high-norm foreground tokens, while VMamba distributes it
across spatial locations through token directions. Both patterns appear to originate from VMamba's block structure, which produces background high-norm tokens associated with direction-based encoding. This unified accounts for the cross model representational divergence at the final stage and VMamba's greater sensitivity to supervision change. The practical consequences follow directly. Magnitude-aligned encoding supports efficient aggregation at standard resolution, where a few dominant tokens suffice for image-level prediction. Direction-based encoding becomes advantageous when token count grows or supervision is spatially dense, as directional information aggregates more stably and reorganizes more readily under fine-tuning.

These results identify token magnitude and direction as key design axes for high-resolution visual backbones. Foreground-aligned magnitude regularization may strengthen standard-resolution classification, while angular regularization or token-wise contrastive objectives may improve dense and high-resolution performance.

\paragraph{Limitations.}
The observed differences are attributed to VMamba's block structure, but the precise contributing components remain unclear, as the SSM cannot be isolated from the surrounding architecture without targeted ablations.

{
\small
\bibliographystyle{unsrt}
\bibliography{refs}
}

\clearpage  
\appendix

\section*{Appendix}


Additional details and results from the different sections are included below.

\section{Experimental Setting}
\label{app:setting}
All experiments use ImageNet-1K~\citep{deng2009imagenet} pretrained weights of VMamba-T and MambaOut-T and are conducted on 4 NVIDIA RTX PRO 6000 GPUs. Unless otherwise stated, all figures and analyses are produced using the ImageNet-1K validation set. We fine-tune the \(224\times224\)-pretrained models on ImageNet-1K at \(768\times768\) resolution for 30 epochs, reducing the original learning rate by a factor of $1/10$ for stable fine-tuning. For semantic segmentation, we initialize from the ImageNet-1K pretrained backbones and fine-tune with UperNet~\citep{xiao2018upernet} on ADE20K~\citep{zhou2017ade} for 160K iterations with batch size 16 and AdamW optimizer~\citep{loshchilov2018adamw}, following the standard protocol used by both VMamba~\citep{liu2024vmamba} and MambaOut~\citep{yu2025mambaout}. For Pascal VOC 2012~\citep{everingham2010pascal}, we train with a batch size of 16 for 20K iterations.

\section{Additional Qualitative Results}
\label{app:qualitative}
\begin{figure}[b]
  \centering
  \includegraphics[width=1.0\linewidth]{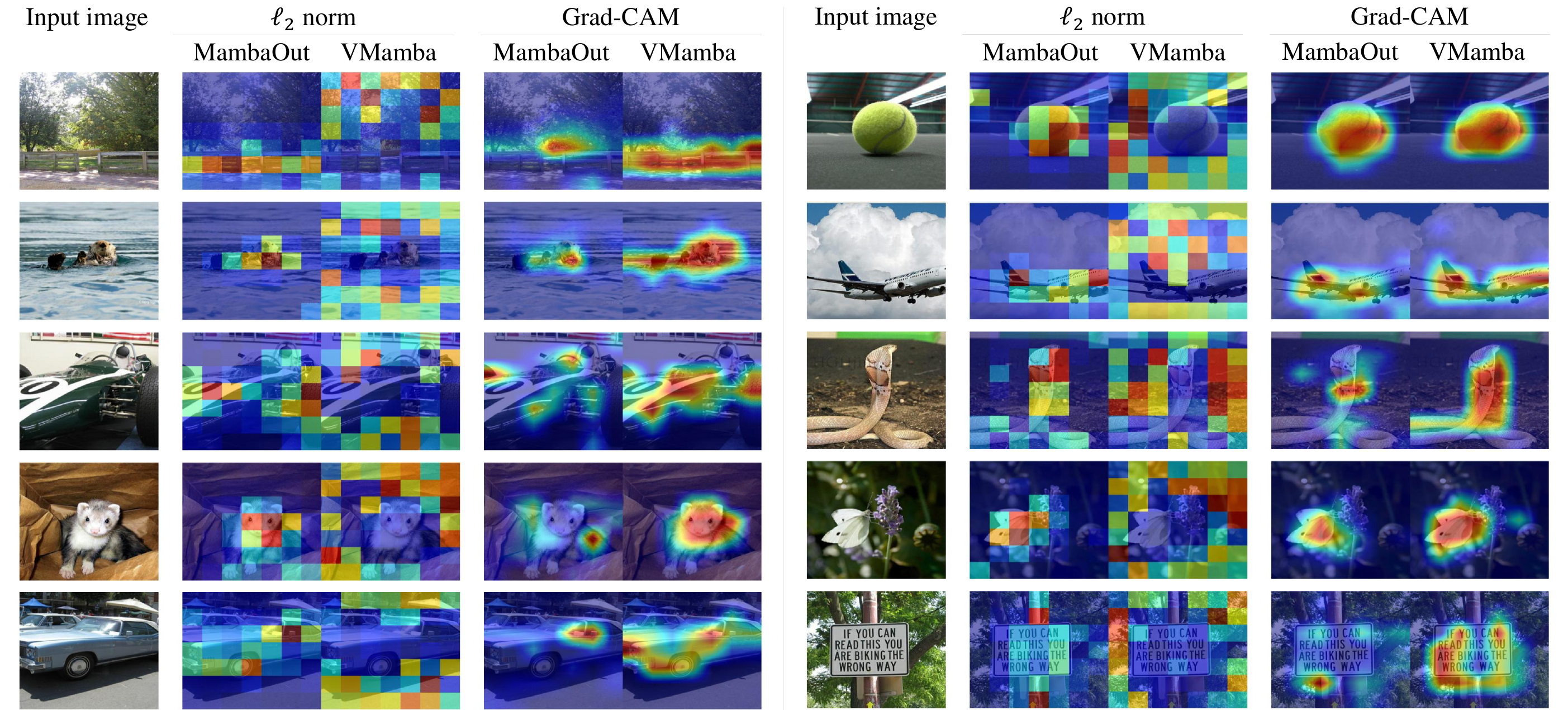}
  \caption{Visual comparison of token \(\ell_2\) norm and Grad-CAM heatmaps from the final block outputs of VMamba and MambaOut.}
  \label{fig:l2norm-app}
\end{figure}

\begin{figure}[t]
  \centering
  \includegraphics[width=1.0\linewidth]{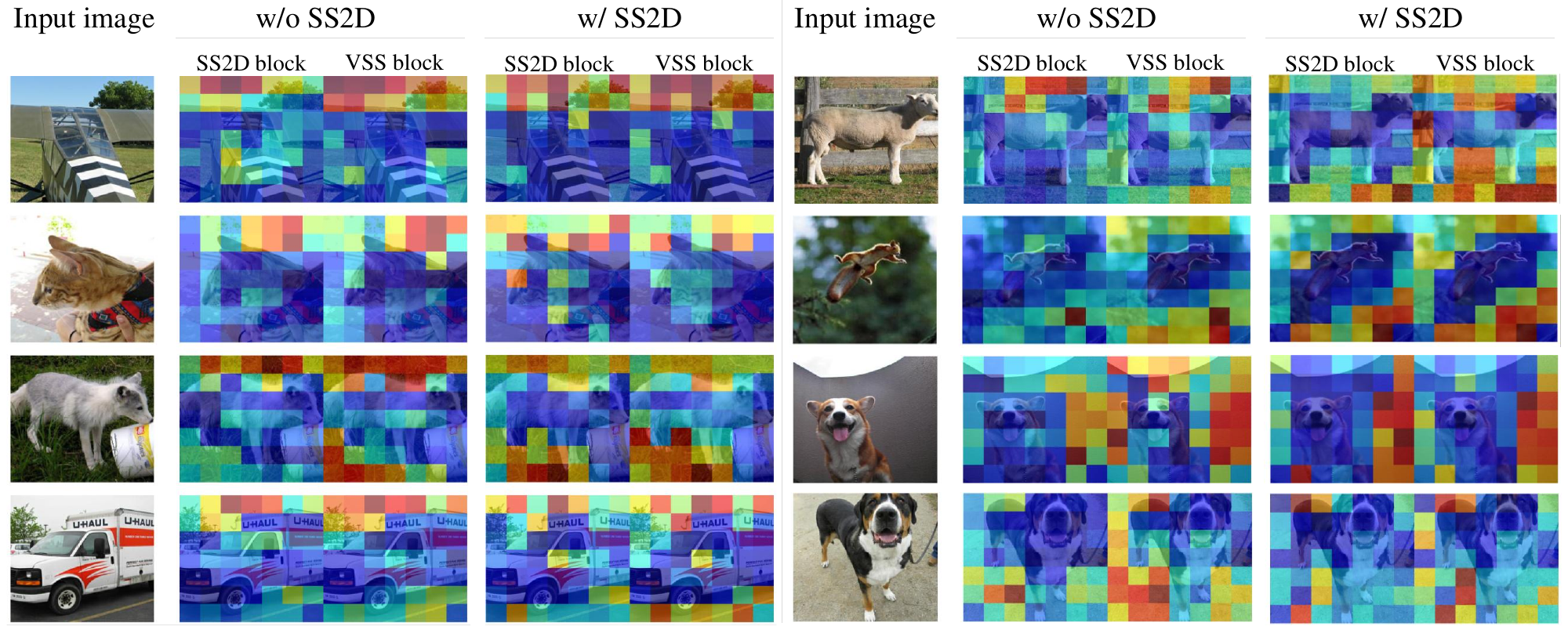}
  \caption{Token $\ell_2$ norm heatmaps at two intermediate outputs (SS2D block and VSS block) under two conditions (with and without SS2D).}
  \label{fig:wossm-app}
\end{figure}

\begin{figure}[t]
  \centering
  \includegraphics[width=1.0\linewidth]{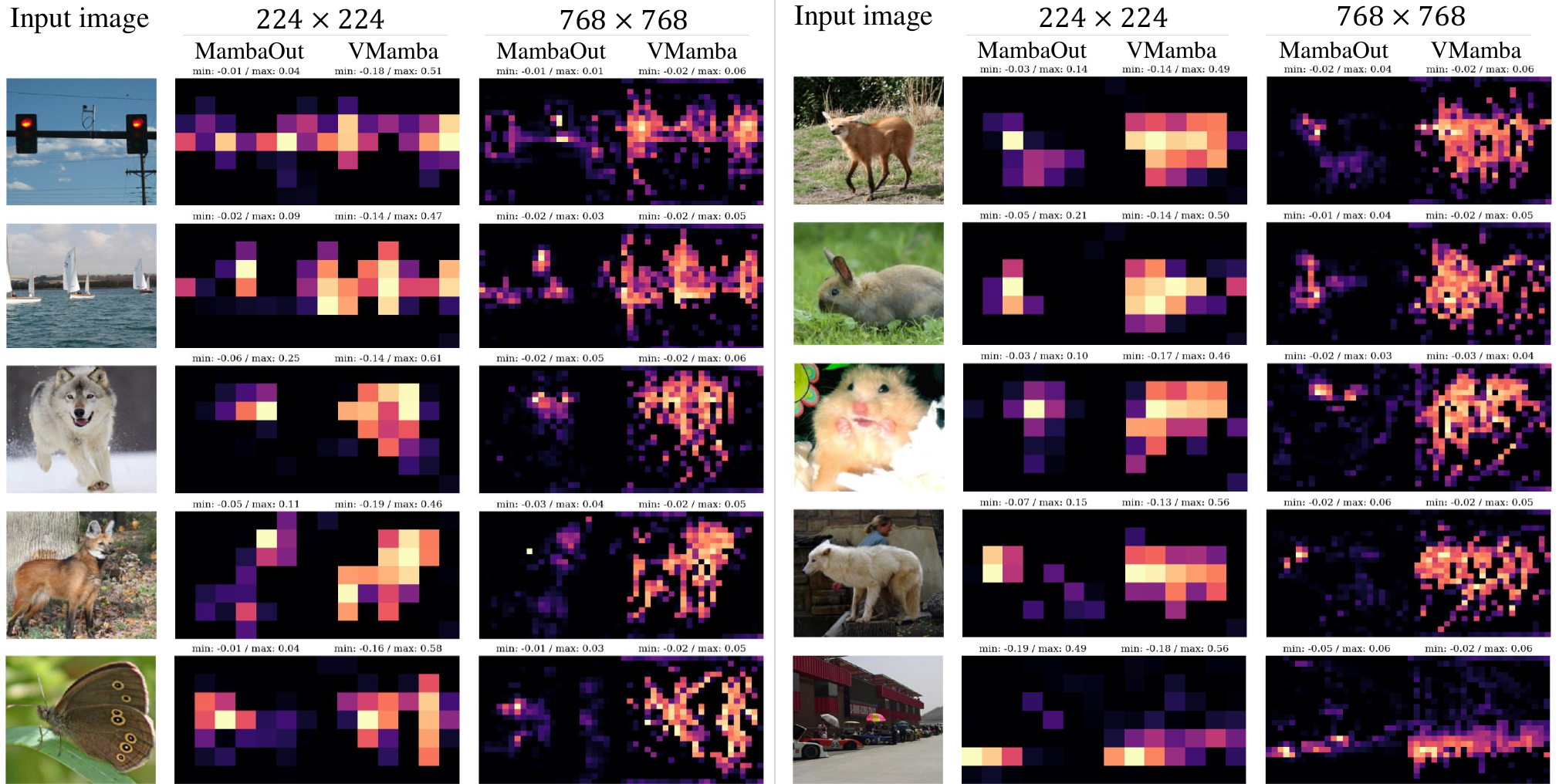}
  \caption{Token replacement attribution at \(224\times224\) and \(768\times768\) resolution for VMamba and MambaOut. For each final block spatial token, we replace only that token with the image-wise mean token and measure the drop in the target class logit.}
  \label{fig:replacement-app}
\end{figure}

Figures~\ref{fig:l2norm-app}, ~\ref{fig:wossm-app}, and~\ref{fig:replacement-app} provide additional qualitative results extending Figures~\ref{fig:l2-grad}, ~\ref{fig:wossm}, and~\ref{fig:token-replace} in the main paper, respectively.

\section{CKA Implementation Details}
\label{app:cka}
In Sections~\ref{sec:cka} and~\ref{sec:spatial_cka}, we measure representational similarity using Centered Kernel Alignment (CKA)~\citep{cortes2012alcka, kornblith2019similarity}, which enables consistent comparison of representations across layers and models. Let $X \in \mathbb{R}^{m \times d_1}$ and $Y \in \mathbb{R}^{m \times d_2}$ denote activation matrices from two layers evaluated on the same set of $m$ samples. The corresponding linear Gram matrices are $K = XX^\top$ and $L = YY^\top$.

Using the centering matrix $H = I_m - \frac{1}{m}\mathbf{1}\mathbf{1}^\top$, we compute centered Gram matrices $K_c = HKH$ and $L_c = HLH$. CKA is then defined as
\begin{equation}
    \mathrm{CKA}(K, L) = \frac{\mathrm{HSIC}(K_c, L_c)}
    {\sqrt{\mathrm{HSIC}(K_c, K_c)\,\mathrm{HSIC}(L_c, L_c)}},
\end{equation}
where the Hilbert-Schmidt Independence Criterion (HSIC)~\citep{gretton2007kernel, kornblith2019similarity} is computed as
\begin{equation}
    \mathrm{HSIC}(K_c, L_c) = \frac{1}{(m-1)^2} \mathrm{tr}(K_c L_c).
\end{equation}
This formulation is invariant to orthogonal transformations and isotropic scaling of representations, enabling meaningful comparison across layers. We compute CKA using the minibatch estimator of~\citep{nguyen2021do}. We sample 2,560 images from the ImageNet-1K validation set per run (batch size 128) and report averages over 10 independent repetitions. This configuration yields results comparable to larger sample settings used in prior work~\citep{raghu2021vision}.

\paragraph{Cross model comparison.}
Following~\citep{raghu2021vision}, we compute CKA between all pairs of blocks in VMamba and MambaOut, producing a cross-model similarity matrix that captures representational correspondence across architectures. For each block, we use the block outputs. Activations are extracted via forward hooks and reshaped to $\mathbb{R}^{m \times d}$ prior to computing Gram matrices. In addition, we compute within model CKA between classification pretrained and segmentation fine-tuned models to quantify representation shifts under different supervision.

\paragraph{Spatial localization.}
Following~\citep{raghu2021vision}, we evaluate spatial localization by measuring the correspondence between token representations and their associated input spatial positions. For each token, we compute its CKA similarity with patch embeddings at all spatial locations, forming a spatial similarity map. A peaked response at the token’s corresponding location indicates strong spatial localization, whereas a more uniform response suggests globally distributed or non-localized representations.

\section{Linear Probing Details}
\label{app:linear_prob}
In Section~\ref{sec:high_norm_linear_probing}, we evaluate the class-discriminability of intermediate representations via linear probing, following the regularized least-squares protocol of~\citep{dosovitskiy2021vit, raghu2021vision}. Let $X \in \mathbb{R}^{n \times d}$ denote the feature matrix aggregated over $n$ samples and $Y \in \{-1, +1\}^{n \times C}$ the corresponding target matrix, where $C = 1{,}000$ is the number of ImageNet classes. The probe weights $W \in \mathbb{R}^{d \times C}$ are obtained by solving
\begin{equation}
    W^* = \arg\min_{W} \;\|XW - Y\|_F^2 + \lambda\|W\|_F^2
        = \bigl(X^\top X + \lambda I\bigr)^{-1} X^\top Y,
\end{equation}
and the solution is recovered in closed form without updating the backbone.

\paragraph{Top-norm token selection.}
We restrict the probe to spatial tokens with high activation magnitude. For each image, we rank all $T$ tokens from the final block by their $\ell_2$ norm and retain the top-$k$\%, yielding $\lceil T \cdot k/100 \rceil$ tokens per image. This focuses the probe on strongly activated regions while discarding low-activation background tokens. We use $k \in \{5, 10, 20\}$ and operate on raw activations. Performance is reported as the average per-token classification accuracy over the selected tokens.

\paragraph{Training setup.}
We train probes using 10 images per class (10-shot), sampled uniformly from the 
ImageNet-1K training set, with regularization parameter $\lambda = 0.1$. Evaluation 
is performed on the full ImageNet-1K validation set.

\section{Token Replacement Attribution}
\label{app:token_replacement}
In Section~\ref{sec:token_replacement1}, we quantify the spatial importance of tokens in the final block using a token replacement attribution method.

\paragraph{Attribution definition.}
Let $X \in \mathbb{R}^{T \times d}$ denote the spatial token matrix at the output of the final block, and let $f_c(X)$ denote the logit for target class $c$. For each token position $i \in \{1, \ldots, T\}$, we define the attribution score as
\begin{equation}
    e_i = f_c(X) - f_c\!\bigl(X^{i \leftarrow \bar{x}}\bigr),
\end{equation}
where $X^{i \leftarrow \bar{x}}$ replaces the $i$-th token with the image-wise mean token $\bar{x} = \frac{1}{T}\sum_{j=1}^{T} x_j$. A larger $e_i$ indicates greater contribution of token $i$ to the target class.

\paragraph{Implementation details.}
The target class $c$ is defined as the top-1 predicted class of each model, and attribution is computed with respect to the model's own prediction. We evaluate inputs at resolutions of $224\times224$ and $768\times768$.

\section{Evaluation Details}

\subsection{High norm test.}
\label{app:high_norm_test}
In Section~\ref{sec:high_norm_test}, for each image, we rank all $T$ spatial tokens from the final block by their $\ell_2$ norm and select the top-$k$\% highest-norm tokens with $k \in \{5, 10, 20\}$, operating on raw activations. Each
selected token is then passed individually through the pretrained classification head, and top-1 accuracy is averaged over all selected tokens across the ImageNet-1K validation set.

\subsection{Unit vector test.}
\label{app:unit_norm_test}
In Sections~\ref{sec:unit_vector_test1} and~\ref{sec:unit_token_decoder}, both evaluations remove token magnitude by normalizing to unit $\ell_2$ norm, leaving only the directional component. For classification (Section~\ref{sec:unit_vector_test1}), every token in the final block is normalized and passed individually to the pretrained head, and top-1 accuracy is averaged over all $T$ tokens. For segmentation (Section~\ref{sec:unit_token_decoder}), each token $x_i$ in every stage feature map fed to the UperNet decoder~\citep{xiao2018upernet} is replaced by
\begin{equation}
  \tilde{x}_i = \sqrt{C}\,\frac{x_i}{\|x_i\|_2},
\end{equation}
where the $\sqrt{C}$ factor keeps the feature scale comparable to standard-normalized inputs. This substitution is applied to all four stage outputs simultaneously, with backbone weights remaining unchanged.

\clearpage 


\end{document}